\documentclass[12pt]{article}

\usepackage{scicite}

\usepackage{times}

\usepackage{graphicx}
\usepackage{etoolbox}
\makeatletter
\usepackage[labelfont={bf}]{caption}
\patchcmd{\@makecaption}
  {\scshape}
  {}
  {}
  {}

\makeatother

\newenvironment{sciabstract}{%
\begin{quote} \bf}
{\end{quote}}

\title{Grasping and Rolling In-plane Manipulation Using Deployable Tape spring Appendages}

\author
{Gengzhi He,$^{1}$ Curtis Sparks,$^{1}$ Nicholas Gravish$^{1\ast}$\\
\\
\normalsize{$^{1}$Department of Mechanical and Aerospace Engineering, University of California San Diego,}\\
\normalsize{9500 Gilman Dr, La Jolla, CA 92093, USA}\\
\\
\normalsize{$^\ast$To whom correspondence should be addressed; E-mail:  ngravish@eng.ucsd.edu.}
}

\date{}

\begin{document} 


\baselineskip24pt


\maketitle


\begin{sciabstract}
Rigid multi-link robotic arms face a tradeoff between their overall reach distance (the workspace), and how compactly they can be collapsed (the storage volume). 
Increasing the workspace of a robot arm requires longer links, which adds weight to the system and requires a larger storage volume.
However, the tradeoff between workspace and storage volume can be resolved by the use of deployable structures with high extensibility. 
In this work we introduce a bidirectional tape spring based structure that can be stored in a compact state and then extended to perform manipulation tasks, allowing for a large manipulation workspace and low storage volume.
Bidirectional tape springs are demonstrated to have large buckling strength compared to single tape springs, while maintaining the ability to roll into a compact storage volume. 
Two tape spring structures are integrated into a bimanual manipulator robot called GRIP-tape that allows for object Grasping and Rolling In Planar configurations (GRIP). 
In demonstrations we show that the continuum kinematics of the tape springs enable novel manipulation capabilities such as simultaneous translation-rotation and multi-object conveyance.  
Furthermore, the dual mechanical properties of stiffness and softness in the tape springs enables inherent safety from unintended collisions within the workspace and soft-contact with objects.
Our system demonstrates new opportunities for extensible manipulators that may benefit manipulation in remote environments such as space and the deep sea. 
\end{sciabstract}

\section{Introduction}
Robotic manipulators with large reach and small storage volume have significant potential for operation in remote environments such as space and the deep sea.  
However, striking a balance between a large manipulation workspace and small storage volume poses a challenging design problem.
Traditional rigid-link based robot arms are fundamentally limited since their physical volume does not change during operation. 
To circumvent this tradeoff between reach and volume, engineers have looked to deployable structures which can expand dramatically in volume from compact to extended states \cite{Pellegrino_undated-xh, Fenci2017-rl}.
However, current deployable manipulators face a fundamental limitation that the volume changing body and the end-effector are typically separate entities.
The end effector thus adds significant weight to the system, it requires cabling to be routed through the deployed structure, and since gripping can only occur at one location the whole system lacks reconfigurability. 
In this work we will present a manipulator that uses deployable structural elements as manipulation surfaces leading to a lightweight, extensible mechanism that is able to achieve novel manipulation capabilities.

Some of the earliest deployable manipulators can be traced back to the kinematic mechanisms of the industrial age and before \cite{Fenci2017-rl}. 
For example, the scissor mechanism may be the simplest example of a linkage system that can dramatically increase in length through actuation of a single degree of freedom. 
Roboticists have leveraged expandable linkages such as the scissor mechanism to create deployable robot arms with gripping end effectors \cite{Suthar2021-uw, Teshigawara2019-ra}. 
Similarly, concentric tube telescoping structures have also been used to expand robot arms prismatically from a collapsed state \cite{Kemp2022-xy}.
While scissor and telescoping mechanisms are limited to prismatic motion, more complex planar or spatial linkage arrangements can yield other three-dimensional modes of expansion \cite{Ding2013-km, Wei2014-xr, You1997-wz}. 
Additionally, kinematic expansion need not be limited to mechanical linkages.
For example sheets of folded paper using origami designs can also be deployed from flattened states to form elongated structures \cite{Rus2018-mv}.
Origami mechanisms can often be treated similarly to linkage systems, while they present new opportunities for fabrication, control, and stiffness modulation \cite{Suk-Jun2018, Wu2021,pneumaticCubes,Yan2019-pq}.
Deployable linkage systems are predominantly made from rigid structures and thus they have the benefit of predictable kinematics.
However, linkage-based deployable manipulators suffer from the same limitation as traditional robot arms: their physical volume never changes and thus they require careful design to pack the structure into a small volume. 
To circumvent this limitation recent advances in deployable robotics have leveraged soft materials which enable volume change through the system's elasticity or flexibility \cite{Wang2016-gc}. 

Soft robots composed of compliant materials enable adaptation to the environment during locomotion \cite{Boudet2021-qt,Ilievski2011-pz,Greer2020-wm,Marchese2014}, responsive shape change for grasping complex objects \cite{Brown2010-lz,Galloway2016}, and inherent safety when interacting with or around humans \cite{POLYGERINOS2015}. 
A recent design paradigm to make deployable mechanisms in soft robotics is engineered length and volume change of the robot body for robots that ``grow''.
Volumetric growth has been developed in systems such as pneumatic actuators \cite{Talas2020-xb, Hammond2017-bi}, robot skins that enable body eversion \cite{ Hawkes2017,eversionRobot,Greer2019-qm,Pengchun2021,Blumenschein2018}, flexible zipping structures \cite{Kim2023-sd, Collins2016-yb}, or additive manufacturing technology \cite{Sadeghi2017-kz,Del_Dottore2024-zt}.  
Growing robots have thus far shown great promise in deploying over extremely long lengths \cite{Hawkes2017}, moving through challenging substrates such as sand \cite{Naclerio2021-cv} or the body for medical procedures \cite{Blumenschein2020-ms}, and relying on simple off the shelf materials. 
While soft growing robots are capable of significantly more volume expansion compared to linkage based mechanisms, they still suffer from the challenge that they must carry a separate end-effector for manipulation.
This limitation makes cable routing through the expanding body a challenge for soft bodied deployable robots. 

Another category of deployable manipulators are those that use a change in material curvature to collapse/expand and to modulate stiffness. 
Tubular deployable structures which can be flattened and rolled into a tight coil and extended into a strong linear beam are the most common form of these systems \cite{Pellegrino_undated-xh}.
In the flattened state the structure is relatively flexible to bending, however when the cross-section is expanded to a circle (or arc) significant strength is provided by the bending resistance of curved materials. 
Tubular mechanisms often are fabricated from composite material such as carbon fiber, or metals such as steel.
Tubular based deployable structures have a long history stemming back to the early days of space exploration and satellite deployment \cite{Fenci2017-rl, Sickinger2004-qg}. 
These mechanisms have been used as deployable beams \cite{Mallikarachchi2014-gh, Leipold2005-wv, Murphey2010-ln} and reconfigurable truss systems \cite{Seriani2015-fn, Yang2022-zw} for robotic applications in space. 

An exciting recent area of development for deployable manipulation is the use of tape-spring based mechanisms for robotic arms. 
Tape-springs are a special example of a tubular structure that can be easily collapsed flat, they exhibit bistable buckling behavior which can be used for energy storage/release \cite{Seffen1997-us}, and they have anisotropic stiffness for novel passive compliance. 
Recent robots have used tape-spring appendages with claw and adhesive end effectors for mobility, enabling the robots to grasp onto cave walls \cite{Chen2022-sl, Schneider2021-an, Newdick2022-lp}.
Furthermore, robot arms that use two antagonistic tape-springs can form local bends to place grasping anchors in challenging to reach locations \cite{Quan2022-un, Quan2023-ee}. 
The curvature-dependent compliance of tape-springs enables joint-link configurations to be reconfigured in robot arms by pinching locations to flatten the curvature \cite{Godson_Osele2022-uj}.
Lastly, small scale tape-springs are being developed for steering needles through tissue \cite{Abdoun2023-uo}.
A common design paradigm of these recent robots is the use of tape-springs as a structural element of the robot, whereas environment manipulation occurs through special end effectors. 

\begin{figure} 

    \centering

    \includegraphics[width = \textwidth]{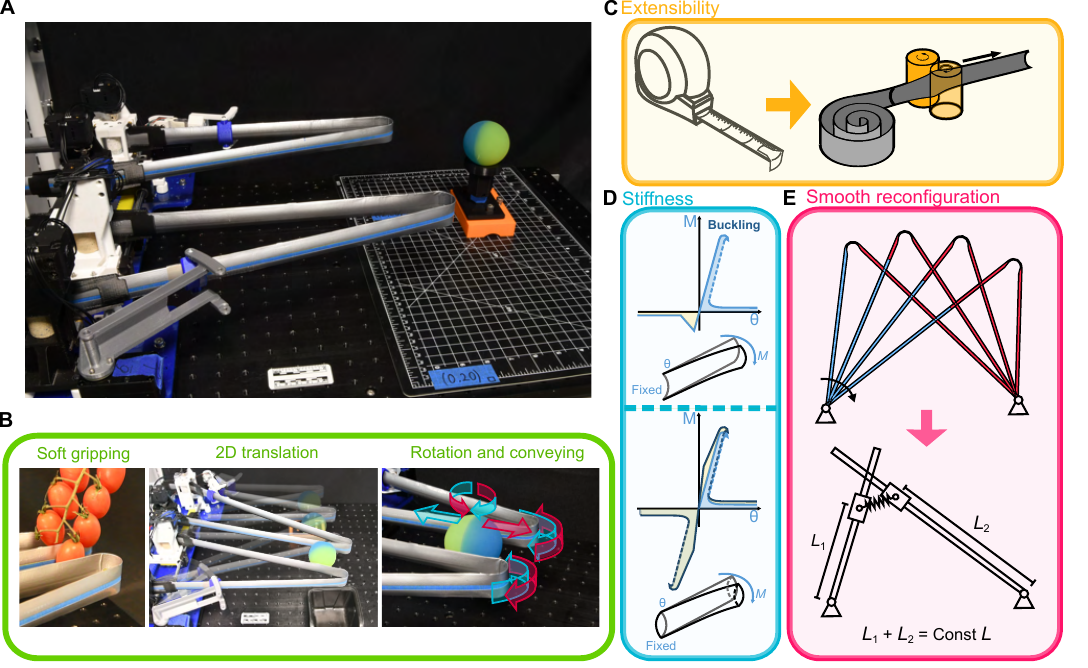}
    \caption{\textbf{Functional basis, implementation, and demonstrated capabilities of the GRIP-tape extensible gripper.} 
    (\textbf{A}) An implementation of two tape spring appendages to form the GRIP-tape two-digit manipulator.
    (\textbf{B}) Capabilities of the tape spring gripper include the ability to interact with soft objects, translate objects over large distances in a 2D plane, and in-grasp manipulation including rolling and conveying objects. 
    (\textbf{C}) Tape spring beams are capable of being rolled into compact spaces and extended over long-distances.
    (\textbf{D}) The beam stiffness is asymmetric in the case of unidirectional tape springs, and symmetric in bidirectional tape springs.
    (\textbf{E}) By bending the tape spring, a reconfigurable appendage is formed. The kinematics of this appendage are modeled as two rotation-prismatic joints coupled through an elastic spring. 
    }
    \label{fig:intro_figure}
\end{figure}

In this work we present a new concept for deployable manipulation using tape-springs in which the deployable structure is the manipulator surface. 
With two-appendages the manipulator robot is capable of object Grasping and Rolling In Planar configurations and we call this robot GRIP-tape (Fig.~\ref{fig:intro_figure}A).
Using the entire length of the tape-spring as a gripping surface makes the arms lightweight (no extra motors or mechanisms needed at the end), eliminates any cabling requirements for an end effector, and enables new novel manipulation kinematics such as multi-object conveyance and rotation (Fig.~\ref{fig:intro_figure}B). 
Furthermore, by developing laminated bidirectional tape-springs GRIP-tape has isotropic bending stiffness in the straight state (Fig.~\ref{fig:intro_figure}C), whereas in the bent state the tape-springs provide high compliance enabling reconfiguration (Fig.~\ref{fig:intro_figure}D, E and movie S1) and the ability to interact with delicate objects.
In the following sections of this paper we will elaborate the design principles for GRIP-tape, and demonstrate how using tape-springs as deployable manipulators can enable novel capabilities unrivaled by other systems.

\section{Results}
\subsection{Design}

\subsubsection{Development of bidirectional tape-spring appendages}
The GRIP-tape robot relies on the mechanical properties of curved beams for gripping stiffness and smooth hinge reconfiguration (Fig.~\ref{fig:intro_figure}D, E).
To begin our design of a tape spring gripper, we first sought to analyze the mechanical performance of tape springs to validate that sufficient gripping force could be supported by deployed tape appendages.

\paragraph{Tape configurations}
A single tape spring like that found in a tape measure is relatively stiff when loaded by a transverse force pointing into the concave direction of the tape curvature (See movie S1). 
However, when loaded into the convex direction of curvature, or with torsional moments along longitudinal axis of the tape, a single tape spring will quickly bend and not be capable of supporting load.
To avoid the anisotropic buckling associated with transverse loading or the twisting failure from torsional loading, we sought to make a symmetric tape spring structure by binding two tape springs together along their length.
This ``bidirectional'' tape design enables the tapes to spool into a compact shape for deployable applications while still retaining the stiffness required for gripping applications. 
Symmetric shell designs have been previously explored for deployable composite booms \cite{Fernandez2017-by}.
We fabricated bidirectional tape springs for GRIP-tape by first aligning two spring-steel tape springs collected from an off the shelf tape measure (Fig.~2A). 
Next we applied an adhesive layer (duct tape) along the sides of the two tapes to adhere them together.

\paragraph{3-point bend tests}
To compare the strength performance of the bidirectional tape with normal tape springs, we conducted a 3-point bending test (See movie S1).
The results, as depicted in Fig.~2B, reveal that compared with the 2 configurations of unidirectional tape, the bidirectional tape exhibits the highest buckling force, indicated by the load peak. 
In the case of the unidirectional tape loaded on the concave side (the green line), the load-displacement curve is smooth instead of a drastic drop, implying the flatting of the curve is smooth rather than buckling at a force threshold.
Due to the geometric relationship inherent in bidirectional tapes, the displacement can be viewed as the combined effect of two unidirectional tapes reacting to the same applied load. 
In tests involving the unidirectional tape loaded on either the convex or concave sides, it is likely that friction between the test platform and the two edges of the tape due to the flatting of the curve hinders deformation, resulting in an undesired increase in the recorded load. 
While in the case of the bidirectional tape, the expansion happens between the two tapes simultaneously, and the contact surface and the test platform has almost no displacement thus the friction is reduced. 


\paragraph{Bend test with rotational base}
We next measured the buckling force of tapes at various lengths to determine the grip force capabilities of extended tapes. 
The experiment consisted of rotating an extended tape spring at the base and measuring the contact force with an an object at a prescribed distance from the base until the tape buckles (Fig.~2C). 
The load cell measures the maximum force that the tape springs exhibit before buckling.

We ran the test on 3 different configurations of double-layered tape springs:  the bidirectional tape and double-stacked tape (two pieces of tape springs stacked in the same direction) loaded in each direction. 
Additionally, a unidirectional tape loaded on its stronger side is also included for comparison.
Note that in this test the bidirectional tape was made with a different fabrication method that held the tapes together in a heat-sealed fabric pouch instead of a duct tape layer. 
In this fabrication, the tapes are free to move relative to each other and do not have the added stiffness of the duct tape to avoid the influence of the duct tape layer.
For each test, the load cell was placed at a distance $L$ from the rotating base and the tape was rotated into the load cell until it buckled. 
The peak buckling force was then recorded and the experiment was repeated three times for each length tested (Fig. 2D). 
The buckling moment of a deployed tape spring is approximately a constant value, thus the force and length of the tape can be fitted to an inversely proportional relationship with an offset (See materials and methods).
The buckling force of the double-stacked tape is almost double the bucking force of the unidirectional tape. 
However, the performance of the bidirectional tape is better than the average of the two double-stacked configurations. 
\paragraph{Fatigue and max buckling force}
An important feature for any robot is the capability to durably operate over long periods of time.
This is potentially problematic for soft robots and deployable systems where the structures of the robot will undergo large deformation and strain.
Specifically for the GRIP-tape robot we require that the tape spring appendages roll and unroll repeatedly during operation without fatigue, and not exhibit failure from errant collisions that cause them to buckle. 
To test the fatigue performance of our bidirectional laminated tape springs we performed a test to measured the buckling force and angle over repeated loading. 
Using the same setup as the buckling test, we mounted a load cell on a track 20 cm away from the starting point of the tape spring. 
During the rotation, the tape measure contacts the load cell, buckles, stops at 16.5 degrees and rotates back to the initial position. 
We gathered the data of 4000 cycles from the load cell measuring the maximum forces before buckling (Fig. 2E), and the angle at which the tape spring buckles (Fig. 2F).
After 4000 trials, the bucking angle and buckling force decreased minimally. 
The buckling angle reduced from 1.28 degrees to 1.16 degrees and the buckling force reduced from 4.97N to 4.915N. 
In the size and force range of GRIP-tape, such an amount of variation is acceptable. 

\subsubsection{Kinematics and control}

\paragraph{GRIP-tape appendages}

We designed GRIP-tape to use two triangular shaped appendages for the left and right gripping surfaces (Fig.~3A).
The straight sections of the appendages act as structural elements that are capable of supporting transverse and compressive loads, whereas the buckled end of the appendage is a rolling hinge that can change the overall shape of the appendage.
The inner sides of the two appendages are the gripping surfaces and our design facilitates object grasping anywhere along the inner length of the appendages.
Each appendage has three independent control inputs: control of the outer beam angle ($\theta_1$), and control of the length change from either the outer or inner side of the appendage ($\delta L_1$ and $\delta L_2$ respectively). 
The inner beam angle is free to rotate and is determined by the total tape length $L$ and outer angle $\theta_1$. 
Both left and right appendages have control over $\theta_1$ and tape length change and thus can be controlled individually. 
Lastly, a single motor controls the symmetric gap width between the inner tapes $w$. 
The combination of three independent control inputs on the left and right appendage, and the width control yields seven overall control variables to position the tapes.  

Individual actuation of the appendage control inputs leads to four primary modes of appendage shape control (Fig.~3B). 
By changing the length of either the inner or outer section ($\delta L_1 \neq 0$ or $\delta L_2 \neq 0$) the outer beam will remain at angle $\theta_1$ while the overall tape length shortens causing the inner beam to bend. 
Note that the $\delta L_1$ and $\delta L_2$ actuators only control the relative change in tape length, while the respective side lengths of the the triangular shaped appendage are determined by the overall tape length and side angle $\theta_1$.
By changing the outer section angle $\theta_1$ the appendage traces a sweeping motion across the workspace. 
If the overall tape length is held constant then the tip will trace out an elliptical shape. 
Changing the tape width $w$ results in a change of the inner section angle and side length while the outer tape remains at a fixed angle $\theta_1$.
Lastly, an equal rate of tape unspooling and retraction on the inner and outer sections ($\delta L_1 = -\delta L_2$) results in a conveyance motion where the overall shape of the appendage is unchanged but the surface motion of the gripper will either move inwards or outwards depending.


\paragraph{Mechanism design and workspace}
The GRIP-tape robot is actuated by a total of seven independent motors as depicted in Fig.~3C. 
    For each appendage, two motors are dedicated to controlling the length change of the inner and outer tape spring beams, and one motor controls the outer angle of each appendage. 
The exact layout and dimensions are presented in materials and methods and Fig. S1A.
The last motor is employed for adjusting the inner width between the appendages, $w$. 
This adjustment is achieved through a rack and pinion mechanism (refer to materials and methods and Fig. S1B).

One tape extruder assembly consists of two 3D-printed cases, each housing a roller. 
One of the rollers is driven by a servo motor and is equipped with sandpaper to enhance friction, while the other roller is passive and freely rotates. 
The rollers are pressed tightly together and the bidirectional tape is passed between them. Supporting guides hold the tapes in position on both sides of the rollers (refer to materials and methods, Fig. S1C and Fig. S1D). 
For the length of an appendage, both extruders are only capable of controlling the total length $L$, and the overall appendage's length (shape) is also influenced by the distance between the 2 extruders $a$ and the orientation of the angular control beam $\theta_4$ (refer to Fig.~3D). 
The appendage's angular orientation is governed by a guiding ring covered by low-friction material and an angular control beam located on each side (refer to materials and methods and Fig. S1E). 
The pivot axis of the angular control beam is affixed to a servo motor mounted on the base and so the outer mounting geometry of the appendages (lengths $c$ and $d$ in Fig.~3D) are predetermined and not adjustable. 
Conversely, the triangle base-width (parameter $a$ in Fig.~3D) for each appendage can be changed and thus is designed to facilitate the grip and conveyance of objects of varying sizes. 
Notably, because the racks are mounted parallel to the x-axis, parameter $b$ is set and not adjustable.


The gripping workspace is determined by the combined range of motion of the appendages, each with an angle restricted annular reach (Fig.~4A). 
For the left appendage, the left angle boundary is constrained by the tape coming into contact with the pivot of the angular control beam. 
The right boundary is limited by the angular control beam colliding with the extruders. 
In the radial direction, the inner workspace radius is defined by the minimum allowable length of the appendage to ensure that the tip does not interfere with the angular control beam.
The outer workspace radius is determined by the allowable extension distance of each tape before they buckle under their own weight.
The right appendage workspace is the mirror of the left and the total workspace is the inclusive combination of the left-right.

Figure 4b shows an image of the total workspace of the gripper. 
We generated a heat map illustrating the maximum grip force that the GRIP-tape can sustain at different locations (Fig. 4B). 
The maximum gripping force was calculated by combining the data from the bend test with rotational base and the workspace analysis. 
The calculation of the maximum gripping force involves the relationship between the maximum force resisting bending and the deployed length (i.e. the distance between the guiding ring where the buckling of appendages happens and the rolling joint) of the supporting side of the appendage (Fig. 2D).
%
Although tape appendages are capable of interacting with objects along their inner surfaces, due to the convenience of directly controlling the location of the appendage's tip we use the minimum length of the appendage to calculate gripping force.
This heat map reveals that the GRIP-tape can support the highest forces near the base, particularly in the central region. 
As the distance from the base increases and the location moves laterally away from the central line (i.e. x = 0mm), the applicable gripping force diminishes.


\paragraph{Inverse kinematics and forward kinematics}
To derive the forward and inverse kinematics we assume an appendage is separated into 3 sections: 1-2) Two straight line sections $L_1$, $L_2$, and 3) a constant curvature arc of length $L_3$ that is tangent to both $L_1$ and $L_2$ (Fig.~3D).
Although the deployed beams of tape still experience some deformation when load is applied, to simplify the model we assume they are rigid links. 
In our design, $b$, $c$, $d$, and $L_4$ are known and constant variables. Since $r$ reflects the load on the appendage, in a scenario with almost 0 load, we use a constant $r = 15~$mm which is determined from measurements. 

By inputting $(x,y)$ and $a$, the inverse kinematics solves for the length of each section $L_1$, $L_2$, $L_3$, and $\theta_4$. 
The inverse kinematics purely involves geometric calculation without solving nonlinear equations. The main steps of inverse kinematics involve determining lengths and angles of the supporting contact section $L_1$, $L_2$, $\theta_1$, and $\theta_2$ with the desired $x,y$ position as the first steps. 
With the relationship $\theta_1-\theta_3=-\theta_2$ and known variable $r$, the length of bending section $L_3$ is derived. 
Within the workspace, $\theta_4$ can be bijectively mapped from $\theta_1$.

The forward kinematics use inputs of total length $L$, arm angle $\theta_4$, and the horizontal distance between the extruders $a$ to solve for the location of the center of the rolling joint $(x,y)$. 
The location of the outer extruder is set as the origin.
The location of the guiding ring on the angle control arm relative to the origin is
$\textit{\textbf{v}} = \left[\begin{array}{@{}c@{}}-d \\c \end{array} \right]+
L_4\left[\begin{array}{@{}c@{}}\cos{\theta_4} \\\sin{\theta_4} \end{array} \right]$. The appendage angle ($\theta_1$) is determined by the unit vector of the guiding ring position $\hat{\textit{\textbf{v}}}=\textit{\textbf{v}}/|{\textit{\textbf{v}}}|$.
The center of the rolling joint can be found through a vector sum of the outer appendage arm and a perpendicular radius of the curve $r$; $\textit{\textbf{X}}=L_1\hat{\textit{\textbf{v}}}+r\left[\matrix{0&1\cr-1&0\cr}\right]\hat{\textit{\textbf{v}}}$. The length of the outer appendage arm $L_1$ is found by solving 3 nonlinear constraint equations. Two equations come from the $x$ and $y$ vector sum of the two appendage arms and their perpendicular radius, as both arms of the appendage must meet at the center of the rolling joint. The final equation is the total length constraint, as the total length of the appendage $L$ is known.
\begin{center}$L = L_1+L_2+r(\theta_1-\theta_2+\pi)$\\
$L_1\sin(\theta_1)+r\sin(\theta_1-\pi/2) =L_2\sin(\theta_2)+r\sin(\theta_2+\pi/2)-b$\\
$L_1\cos(\theta_1)+r\cos(\theta_1-\pi/2) = L_2\cos(\theta_2)+r\cos(\theta_2+\pi/2)+a$
\end{center}
Both inverse and forward kinematics are presented in materials and methods and Fig. S2

The accuracy of the GRIP-tape appendage control was evaluated using a motion capture software OptiTrack Motive and OptiTrack Primex 13 cameras. 
To measure the position of the appendage arc within the workspace.
In the first experiment we commanded the appendage to move to 16 locations across a 4 by 4 grid over the workspace (Fig. 4C), with 6 trials for each location.    
Across the 6 trials for each of the 16 target points norm of the vector distance between the desired x-y location and the measured x-y location was determined to be an average position error of 3.72$\pm$0.35~mm. 
To demonstrate the motion control capabilities from the GRIP-tape kinematics we conducted a second set of experiments in which we moved the appendage tip along a variety of path shapes (Fig. 4D).
Path following was tested through three cycles of path following across four polygonal shapes.
The motion paths were repeated consecutively without any calibration between.
We observed good tracking behavior between the desired and observed paths (Fig. 4D and movie. S3) indicating that GRIP-tape is capable of suitable positioning control.

\subsubsection{Appendage mechanics}
\paragraph{Bidirectional tape extensibility}
In each appendage of the GRIP-tape, two tape extruders adjust the length of the appendage. 
By extending or retracting tape with the extruders, the overall length and reach of the appendage can be changed.  
To determine how far an appendage can reach, the GRIP-tape was mounted horizontally and a tape appendage was extended parallel to the floor until it buckled under its own weight.
This was done with both a unidirectional tape appendage and a bidirectional tape appendage. 
The longest successful extension of a unidirectional tape appendage reached was 3.5ft.
Due to gravity, oscillation, imperfections in fabrication, and the anisotropy of the unidirectional tape; it buckles on itself and collapses. 
A bend is formed diagonally across the tape spring that causes the end of the appendage to hang down towards the floor. 
The bidirectional tape does not have the same anisotropy, so it performed much better. 
The bidirectional tape successfully extended to a max length of 5 ft. Unlike the unidirectional tape, the bidirectional tape is limited primarily by weight and the stiffness of 3d printed parts. 
As the bidirectional appendage reached out farther, it began to sag under its own weight and caused bending in the base and extruder assembly as well. 

To reach high extensions of the appendage while maintaining a small size, it is necessary to store long lengths of the tape springs in a small volume.
When used in a tape measure, tape springs are stored in a compact form by winding them on a spring-loaded spool. 
This can also be done for the unidirectional tape by wrapping both sides of the appendage tape around a spool on the opposite side of the extruders.
However, bidirectional tapes are not capable of this. 
The two tapes inside the wrapping are unable to slide (shear) relative to each other. 
When the tape is wrapped around a spool, there is a mismatch in the length of material needed for the tape on the inside of the coil and the tape on the outside of the coil. 
This causes the bidirectional tape to form bumps and kinks as the length mismatch builds up. 
These imperfections prevent even coiling of the tape. 
To address this issue, our solution is to bind the two tapes with a low-friction sleeve between the adhesive and the inner tapes. 
Since the adhesive is not directly attached to the tapes, they can slide internally with respect to each other and prevent the problematic accumulation of shear strain along the tape length (movie S2). 
By adding an external sleeve to the bidirectional tape fabrication process we are able to spool up the appendage tape and unroll it during deployment.

\paragraph{Soft-contact object interaction through hinge serial compliance}
The structure of one appendage of the GRIP-tape can be represented kinematically as two prismatic links, with constant length constraint, that are connected by a nonlinear spring at the tip of the appendage (Fig. 1E). 
As GRIP-tape interacts with an object the bend curvature at the end of the tape appendage undergoes changes corresponding to the change in gripping force. 
As the gripping force increases, the radius of curvature decreases, leading to a spring-like resistance. 
To quantify this resistance, different configurations of tape were bent 180 degrees, and the tapes were subjected to compression tests. 
The force readings during compressing the tape were recorded as a function of displacement allowing for us to measure the ``stiffness'' of tape contact mechanics (Fig.~5A).
After fitting the compression data to a polynomial curve, the relationship between loading and deformation of the spring-like bending is derived (see materials and methods). 

The non-linearity of the force-displacement relationship indicates that a higher load leads to higher spring stiffness. 
The results for the two unidirectional tapes are similar, for the bidirectional tape, the total load exceeded the sum of the loads of the unidirectional tapes under the same displacement (Fig.~5A). 
This discrepancy may arise from friction between the two pieces of tape and the stretching stiffness of the duct tape. 
It is worth mentioning that due to the properties of the duct tape adhesive that binds the bidirectional tapes, the loading stiffness curve and unloading stiffness curve exhibited a small amount of hysteresis during a loading cycle (Fig.~5B).
The effective elasticity of the tape-spring bend introduces a degree of serial-compliance in the gripper that allows for soft gripping of delicate objects (Fig.~5C). 
As a means of comparison to other elastomeric soft grippers we envisioned that the tape-spring bend joint is a $15$mm$\times 25.4$mm$ \times30$mm sized cube, and calculated an effective Young's modulus of this mechanism.
Comparison of the approximate Young's modulus of the tape-spring bend with other soft materials in Fig.~5D, indicates that the contact interaction mechanics performance are broadly similar to silicone elastomers.

\subsection{Grasping capabilities and demonstrations}

In the prior section we demonstrated the unique mechanics of deployable tapes and our ability to exert simple kinematic control over the appendages.  
We now illustrate the gripping capabilities of the two appendage GRIP-tape system. 

\subsubsection{Gripping, translating, and rotating objects}
Once both appendages of the GRIP-tape contact the target object with a sufficient load, a grasp is formed and we can translate the object by simultaneously moving the left and right appendages to different target positions (refer to materials and methods and Fig. S4). 
In our first demonstration we grasp a rubber ball at the tip of the appendages,  translate the ball to a desired position, rotate the ball in place, and then convey the ball towards the gripper base and release it into a bin (Fig. 6A and movie S3).
The GRIP-tape is capable of grabbing objects of varied shapes and stiffnesses such as an entire tomato vine (movie S4).

In-place, continuous rotation is a unique capability of the GRIP-tape kinematics.
Object rotation is achieved by displacing the object contact points in opposite directions (movie S3). 
The overall length of the appendage can be kept constant during rotation by spooling and unspooling at the base of each appendage. 
Thus, if we want to rotate a ball of a radius $r$ by 90 degrees,
the surface on the contact section of the tape of the left appendage has to extend $\pi r/2$. 
To keep the ball at a constant location, the contact section of the tape of the right appendage has to retract $\pi r/2$. 
Accordingly, the supporting section of the left appendage retracts $\pi r/2$, and the supporting section of the right appendage extends $\pi r/2$. 

In-place rotation can be used to accomplish challenging picking operations such as twisting a tomato off of the vine (Fig.~6B and movie S4). 
The soft-contact capabilities of the tape allow for applying light pressure to the tomato surface, and a continuous in-place rotation twists the tomato releasing it from the vine. 
An additional function of GRIP-tape is the capability to convey objects inwards to the base while keeping the overall shape constant (Fig.~6C and movie S4).
Object conveyance can even be extended to multi-object conveyance where several objects can be grasped and simultaneously translated back to the base of the gripper.
We demonstrate an example of this in Figure~6C in which we convey cherry tomatoes to the base.

In Fig~6D we show a lifting example of grasping a fresh lemon from a distance of approximately 50~cm, vertically lifting the lemon and orienting it over a bin, and conveying the lemon into the bin. 
In movie S4 we show several other examples of object manipulation including grabbing, translating, and screwing in a lightbulb.
All demonstrations are performed through teleoperating the robot with a custom designed app that enables joystick control (movie S4 and Fig. S5).

\subsubsection{Passive compliance and robustness}
The passive compliance of the tape appendage gives it unique capabilities for interacting with objects. 
For example, if there is an obstacle impeding the planned trajectory, the appendage is capable of deforming around the obstacle and continuing on its way to reach the desired ending position (Fig. 7A and movie S5). 
If there is a known obstacle in the way that would prevent the appendage from reaching the target normally, the appendage can deform against the known obstacle and remain capable of grabbing and conveying the target object.
Fig. 7B and movie S5 show an application of object grasping and conveyance while the appendages are deformed from contact with a wall. 

The material properties of spring-steel and the cross sectional curvature of the tape provides tape-springs the robust ability for self-recoverability. 
A tape spring bent out of its original shape will snap back to its original configuration when released.
The GRIP-tape appendges self-recoverability property is demonstrated in movie S5. 
The appendage is struck while trying to complete a task. The strike causes the appendage to buckle. However, the appendage made of bidirectional tape rapidly rebounds and successfully completes the task.

\subsubsection{Force sensing}
To allow for force feedback from the GRIP-tape, a load cell was built into the angular control arm (Fig.~8A). 
Assuming the moments on each section of the appendage are balanced, and knowing $\theta_4$, $a$, and $L$ we can derive $\theta_1$, $\theta_2$, $L_1$, $L_1'$, and $L_2$ by forward kinematics (refer to Fig.~8B). 
The relationship between the force on the contact section and the read of the load cell is:
\begin{center}
$F_2'=\biggl(\Bigl(\bigl(F_{read}L_1'/{\cos{(\theta_1-\theta_4)}}+\tau_1\bigr)/{L_1}\Bigr)L_2+\tau_2\biggr)/L_2'$
\end{center}
$F_{read}$ is the reading of the load cell, and $\tau_1$ and $\tau_2$ represent the internal bending torque of the tape spring. 
To obtain $\tau_1$ and $\tau_2$ at different angles, we measured the internal torque with a bi-directional tape with a pinched point by bending it around this point to a set of different angles. 
By curve-fitting these data with spline (Fig. S6), we developed a relationship between the bending angle and $\tau$. 
To demonstrate the force sensing accuracy we placed another load cell at the tip of the appendage with $L_2'=L_2$ and we compared the force prediction from the base load cell with the actual contact force measured by the contact load cell. 
An example of the comparison between the real force on the tip of the appendage measured by the additional load cell and the calculated value $F_2$ derived from the force reading $F_{read}$ on the base is shown in Fig.~8C demonstrating very good agreement.

\subsubsection{Automatic gripping}
The force feedback in the GRIP-tape allows for locating the position of an object and grasping it without human intervention.
We designed several motion steps to complete automatic gripping (Fig.~8D-I and movie S6).
In step 1 (Fig.~8D) and step 2 (Fig.~8E) we locate the object with the left and right appendages by sweeping first the left appendage inwards until a contact is detected, and then the right appendage is swept inward until contact is detected. 
The steps end when an increase of the force estimate of the $F_2$ is detected with a contact detection threshold value of $0.25$~N. 
After steps 1-2 the left and right appendages are in contact with the object. 
The appendage orientations define a gripping axis along which the object is located.
However, we cannot yet determine the size of the object or the distance of the object along this center line. 
For step 3 (Fig.~8F) and step 4 (Fig.~8G), the GRIP-tape opens its appendages and reconfigures the contact sections to be parallel to the gripping axis and then the tapes are moved closer to each other until contact is once again detected by the force sensor. 
This measurement step now provides the width of the object $w$ which is the distance between the parallel appendages.
However, at this point the location of the object along the length of the gripping axis is not determined. 
Lastly, to determine the object distance along the grip axis we retract the left appendage until contact is lost (measured by a decrease in force from the force sensor). 
Once contact is lost we are able to determine the exact object location in the workspace from forward kinematics (Fig.~8H). 
Once the object location has been determined in the workspace we can move the two appendages to the location of the object, form a grasp, and move the object to the desired location (Fig.~8I and movie. S6).

\subsubsection{Rotation with feedback control}
Force feedback also enables the GRIP-tape system to provide closed-loop contact force control. 
This can be useful for example in cases where a non-round object needs to be rotated. 
Round objects can be rolled in the grip by simply moving the surfaces of the appendage in opposite directions.
The object roundness results in a constant grip force while the object rotates (since the cross-sectional width of the object is not changing between the grip points). 
While this open-loop rotation works well for round objects, objects of unknown or irregular shape pose a challenge for rotation in a grasp. 
The softness of the appendage tip can accommodate some small deviations in grip width variation. 
However, excessive width changes during rotation can lead to either loss of grip (Movie. S7), buckling of the appendage, or slipping of tapes inside the tape extruder due to increased gripping force and excessive friction between the tapes and the guiding ring.

To address this issue, we implemented force feedback from the contact force sensor was used to achieve and maintain a desired gripping force during non-round object rotation. 
The feedback control adjusts the goal configuration of the appendages to keep the desired force contact force constant. 
The force controller is a simple control loop in which the contact force error is used in a proportional controller to servo the grip width. 
If the force reading is less than the desired force, the grip width between the appendage ends is decreased. 
If the force reading is higher than the desired force, the grip width distance between the endpoints is increased. 
Implementing this force feedback control allows for successful, controlled rotation of elliptical and other non-constant grip width objects. 
We compared the performance with and without force feedback by rotating an ellipse and demonstrate that open-loop rotation causes the object to be dropped, while closed-loop force feedback rotation successfully maintains a grasp during rotation.
A demonstration of these capabilities is presented in Fig.~8J and movie S7.

\section{Discussion and conclusion}
In this work we have presented the design and testing of GRIP-tape, a soft robotic gripper constructed from tape springs that is capable of object Grasping and Rolling In Plane (GRIP-tape). 
The GRIP-tape has many of the same advantages as other soft robot grippers, with compliance allowing for interaction with unknown or complex objects \cite{Kim2019-pk}, being inherently safe for human interaction \cite{POLYGERINOS2015}, and being able to reconfigure its own shape \cite{Usevitch2020-sc}. 
The tape appendage mechanism keeps the beneficial properties of robot compliance while also adding additional capabilities for simple in-hand rotation and high extensibility. 
Thanks to the features of tape springs, the inherent maximum load and self-recoverability of tape appendages provide safety during interaction and ensure robustness in gripping  applications.
Owing to the continuity and consistency of the tape appendage, the rolling mechanism allows for the seamless renewal of material if any part is permanently damaged during operation, eliminating the need for complicated manual operations. 
Additionally, due to its geometrically defined shape the tape appendages can be precisely controlled.
These key features make the GRIP-tape a versatile system, well-suited for a variety of applications encompassing object gripping, dual modes of translation/rotation, multi-object conveyance. 

A novel aspect of the GRIP-tape is the ability to rotate objects in place while in a grasp. 
Prior grippers have been developed with similar capabilities however they have relied on the integration of active surfaces such as conveyor belts mounted onto a traditional rigid gripper.  
For example, the underactuated modular finger featuring a pull-in mechanism \cite{Kakogawa2016} and the Sheet-Based Gripper \cite{Morino2020} both used driven belts as the contact surface of each finger, aimed at pulling in and gripping, refining the object-picking process. 
Other examples of active surfaces include Velvet Fingers \cite{Tincani2012,Krug2014-dj} and an active surface gripper consisting of an underactuated finger and rigid thumb \cite{Ma2016-hn}, prioritizing in-hand manipulation and the stability of grasp. 
These advancements leverage the dynamic surfaces on the fingers to enable in-hand maneuvers such as twisting and pulling, thereby enhancing the overall performance during gripping actions.
However, the need for rollers within the fingers or the finger's supportive skeletal structure for the ``conveyor belt'' constrains the choice of rigid materials for constructing the fingers, thereby restricting their flexibility. 
GRIP-tape alleviates these challenges by using the compliant structure of the appendage as the active surface. 

The tape appendage structure also allows for sensing and haptic feedback during use operation. 
We have demonstrated sensing and feedback capabilities using the load cell embedded in the angular control beams.
The load cell can measure the force applied to the appendage and then the applied force can be calculated. 
We can map the force reading of the load cell to the force applied to an object anywhere on the appendage as long as the location of the object is known. 
These capabilities allowed for the development of automatic functions such as searching, measuring (size and location), and gripping using built-in force measurements and structural compliance. 
Furthermore, the applied force can also be estimated by an operator through visual inspection, as the radius of curvature at the ends of the appendage decreases when the gripping force increases. 
This could allow a trained operator to judge at a glance how much force is being applied at any given time while teleoperating the GRIP-tape system.

Potential applications for deployable manipulators include agriculture, space, and sea environements.
Agriculture is an industry where new robotic actuators are highly desired. 
Automation and robotics can help increase productivity and improve quality of life for workers. 
However, agriculture remains a challenging environment for robotics as it is unplanned and variable, requiring a capability for adaptation. 
Additionally, for a robot to be adopted on a farm it must be both low cost and safe around humans \cite{Bechar2016}. 
The GRIP-tape system may find use in agriculture applications where autonomous picking and inspection are required. 
The tape springs that form the actuated appendages are both cheap to produce and safe to use for human interaction due to their buckling properties. 
Furthermore, the different actuation modes of the GRIP-tape could be particularly useful in picking applications; for example reaching out and grabbing a fruit, twisting it off the plant, then retracting and conveying the fruit back to a central body for inspection. 
The compliant and spring-like nature of the end joint would also allow for force control while harvesting produce that would lower the chance of damaging a fruit while harvesting it.

Additionally, the lightweight and high extensibility of the GRIP-tape manipulator could be useful in environments where volume and weight are at a premium. 
Tape springs are already used as for space systems such as antenna booms, satellite solar arrays, and structural supports \cite{Seffen2000,Wilkie2021-oe}.
A tape spring based actuator could be similarly effective, allowing for highly extensible actuators to be sent on space missions with significantly lower weight and volume than rigid link robots with the same reach distance. 
Similarly, an extensible gripper could be useful in deep sea environments where space is limited.
Sea caves are complex and space limited environments that are interesting places to search for new life \cite{Richmond2020}. 
The narrow constraints of the caves limit the equipment that can be carried inside, motivating the use of actuators that can be packed into a small volume and then extended.

Overall, the GRIP-tape is an example of a broader class of soft, curved, reconfigurable, and anisotropic mechanisms (SCRAMs) that provide a broad repertoire of mechanical properties for soft robot development \cite{Jiang2021-ce, Sparks2022-eg, Sharifzadeh2021-ps, Jiang2020-do}. 
The GRIP-tape has been developed with the goal of creating an extensible gripper by utilizing the curvature properties of tape springs. 
Future manipulators based on this concept can be designed for more complex object control include 3D motion and multi-axis object rotation. 
The ease of fabrication and unique mechanical properties of tape springs make them ideal structures for high extensibility manipulators.

\bibliography{sciroedit_v1}
\bibliographystyle{Science}

\section*{Acknowledgments}
Funding support was provided through the Mechanical and Aerospace Engineering Department at UCSD. 
This material was based upon work supported by the National Science Foundation under Grant No. 1935324.  
Any opinions, findings, and conclusions, or recommendations expressed in this material are those of the author(s) and do not necessarily reflect the views of the National Science Foundation.

\section*{Supplementary materials}
Materials and Methods\\
Fig. S1. CAD of the GRIP-tape.\\
Fig. S2. Illustration for the inverse kinematics.\\
Fig. S3. Different transformation modes of the appendage.\\
Fig. S4. Width adjustment on the base during gripping and translation.\\
Fig. S5. App design.\\
Fig. S6. Data points and curve fit of the internal torque of bidirectional tape.\\
Movie S1. Properties of tape spring.\\
Movie S2. Bidirectional tape - extension and spooling.\\
Movie S3. Demonstration of basic abilities.\\
Movie S4. Applications.\\
Movie S5. Features of the tape spring appendages.\\
Movie S6. Automatic gripping.\\
Movie S7. Feedback Control.\\


\clearpage

\clearpage
\begin{figure}
    \centering
    \includegraphics[width = \textwidth]{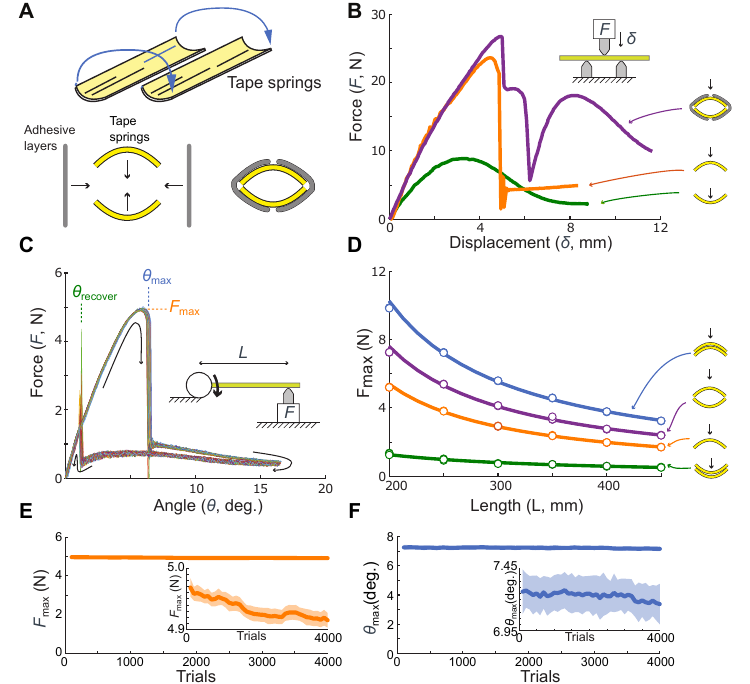}
    \caption{\textbf{The bucking force of different configurations of the tape spring and results of the fatigue test.} 
    (\textbf{A}) Structure of bidirectional tape.
    (\textbf{B}) Results of 3-point bend test of different tapes.
    (\textbf{C}) 4000 cycles of fatigue test.
    (\textbf{D}) Buckling force $F_{max}$ (data and fitted curve) of different setups of tape spring with response to the distance between the fixed point and external force. 
    (\textbf{E}), and (\textbf{F}) Trend of the buckling force and buckling angle.
    }
    \label{fig:experiment1}
\end{figure}

\clearpage
\begin{figure}
    \centering
    \includegraphics[width = \textwidth]{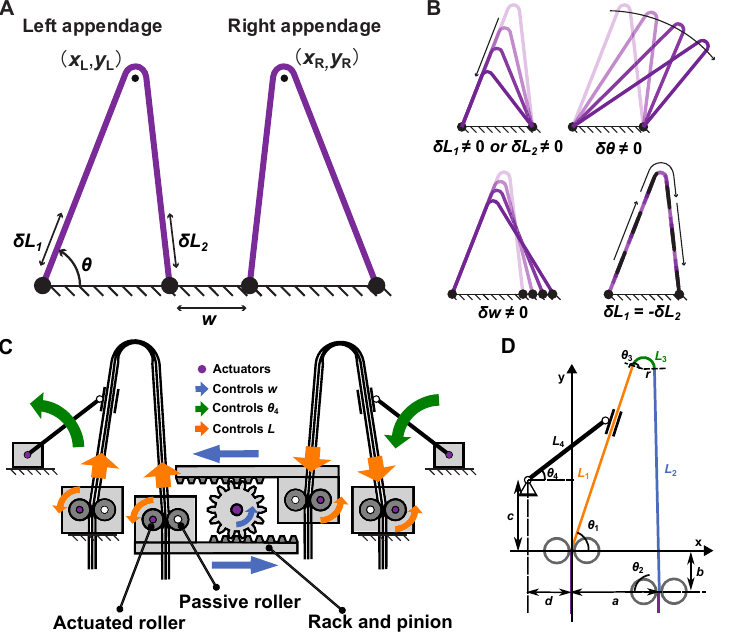}
    \caption{\textbf{Kinematics and design of the GRIP-tape manipulator.}
    (\textbf{A}) GRIP-tape is composed of a left and right digit. 
    Each digit has independent control over the tip location denoted by the black dots. 
    (\textbf{B}) The four insets show the basic modes of appendage control.
    (\textbf{C}) A schematic of the overall control inputs for the two tape spring appendages. 
    (\textbf{D}) Representative model of the left-appendage with actuation inputs from two roller units ($\theta_1$, $\theta_2$) that control the left-right length of the appendage, and a rotational input ($\theta_4$) that controls the angle of the appendage.
    }
    \label{fig: Design}
\end{figure}

\clearpage
\begin{figure}
    \centering
    \includegraphics[width = \textwidth]{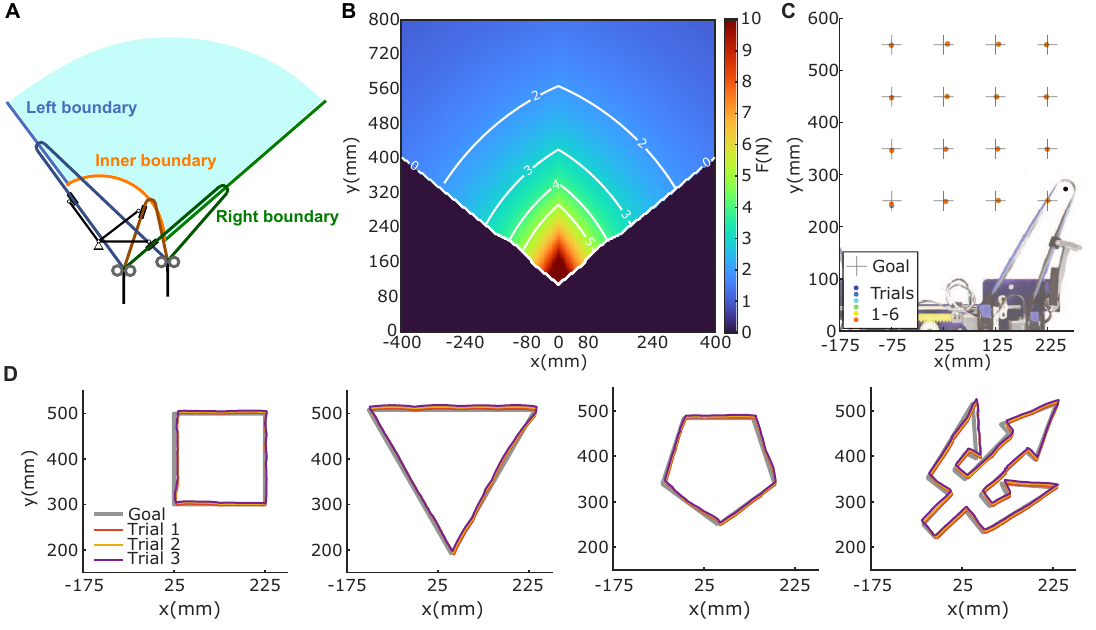}
    \caption{\textbf{Demonstration of GRIP-tape appendage kinematics} 
    (\textbf{A}) The workspace of the left appendage.
    (\textbf{B}) Combined workspace of both appendages with the maximum gripping force computed from buckling measurements and indicated with a color map.
    (\textbf{C}) Inverese kinematics position error of the right digit tip location from over six trials.
    (\textbf{D}) Results from right digit tip location tracing over three trials with four different shapes. See movie S3 for video.
    }
    \label{fig:kinematics}
\end{figure}

\clearpage
\begin{figure}
    \centering
    \includegraphics[width = \textwidth]{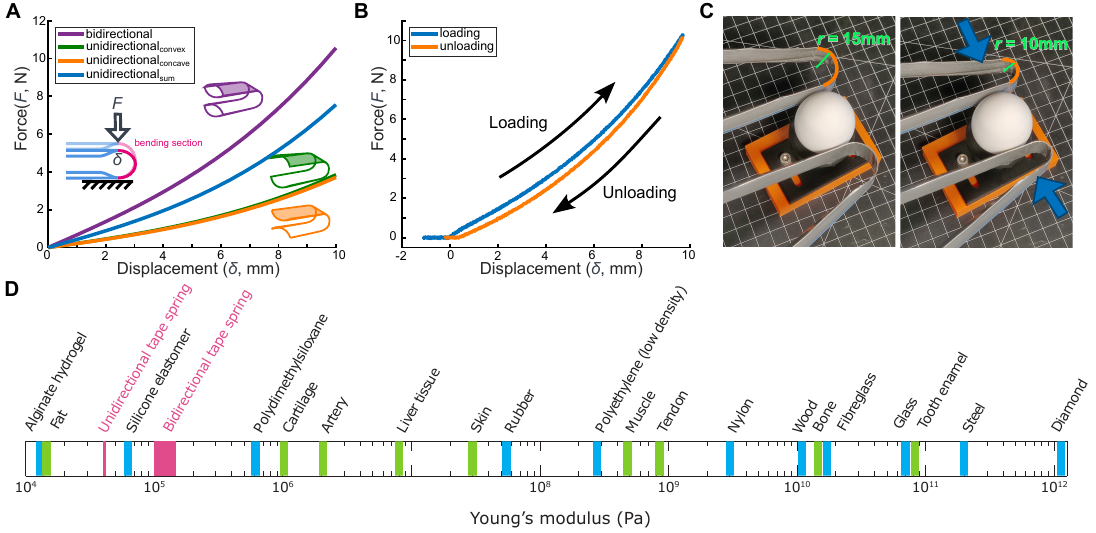}
    \caption{\textbf{Stiffness of the spring-like curve of a bent tape spring.}
    (\textbf{A}) Curve fitting and comparison between the stiffness of bidirectional and unidirectional tapes.
    (\textbf{B}) Raw data of the loading and unloading loop of the bidirectional tape.
    (\textbf{C}) Soft pinch on an egg.
    (\textbf{D}) Young's modulus compared to other materials(Adapted from \cite{Rus2015}).
    }
    \label{fig:tip_stiffness}
\end{figure}

\clearpage
\begin{figure}
    \centering
    \includegraphics[width = \textwidth]{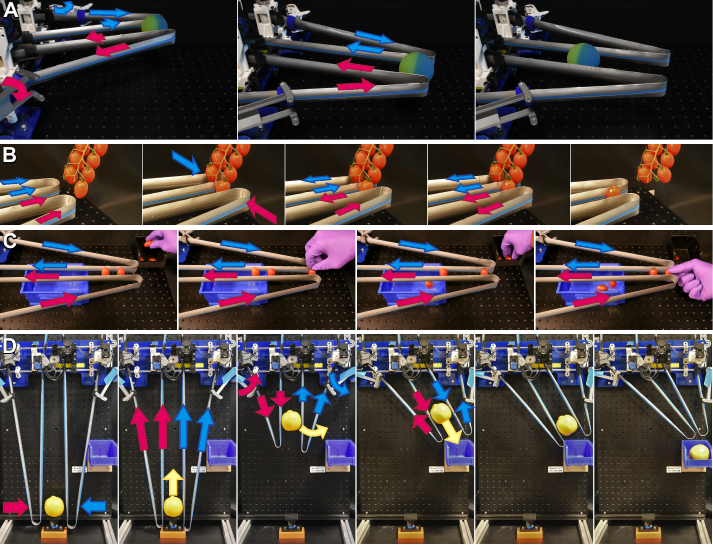}
    \caption{\textbf{Demonstrations of basic applications.} 
    (\textbf{A}) Demonstration of soft gripping and twisting.
    (\textbf{B}) Demonstration of conveyor belt mode.
    (\textbf{C}) Demonstration of lifting.
    (\textbf{D}) Deformation and recovery after hitting an obstacle.
    }
    \label{fig:demo}
\end{figure}

 \clearpage
 \begin{figure}
     \centering
     \includegraphics[width = \textwidth]{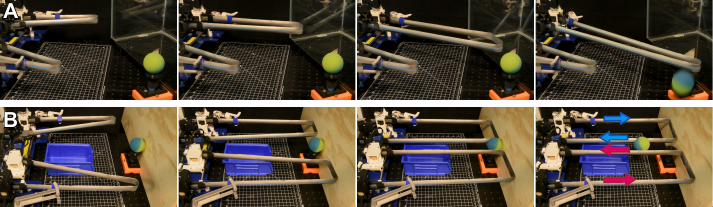}
     \caption{\textbf{Passive compliance.} 
     (\textbf{A}) Deformation and recovery after hitting an obstacle.
     (\textbf{B}) Application of deformed appendages.
     }
     \label{fig:auto}
 \end{figure}

\clearpage
\begin{figure}
    \centering
    \includegraphics[width = \textwidth]{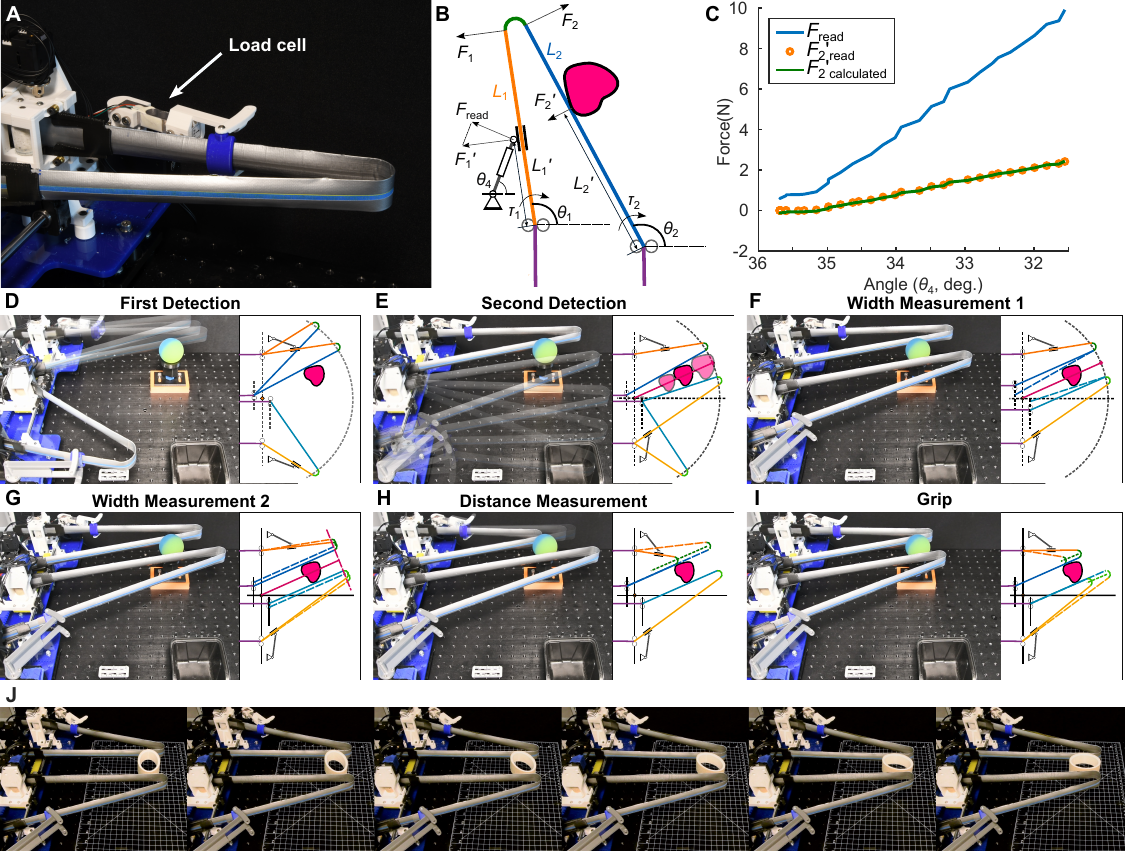}
    \caption{\textbf{Force sensing and automatic searching, testing, and gripping.} 
    (\textbf{A}) Load cell on the angular control beam.
    (\textbf{B}) Balance of moment.
    (\textbf{C}) Comparison of calculated and actual force vs angle of $\theta_4$.
    (\textbf{D}) First detection.
    (\textbf{E}) Second detection.
    (\textbf{F}) and (\textbf{G}) Width measurement.
    (\textbf{H}) Distance measurement.
    (\textbf{I}) Gripping with the tip.
    (\textbf{J}) Object rotation with force feedback.
    }
    \label{fig:autodemo}
\end{figure}


\end{document}